%% file: main.tex
\documentclass[10pt,conference]{IEEEtran}
\usepackage[utf8]{inputenc}
\usepackage{balance}
\usepackage{cite}
\usepackage{amsthm}
\usepackage{multirow}
\usepackage[table,xcdraw]{xcolor}
\usepackage{hyperref}
\usepackage{url}
\usepackage{xspace}
\usepackage{graphicx}
\usepackage{amsfonts}
\usepackage{amsmath,amssymb}
\usepackage{textcomp}
\usepackage{acro}
\usepackage{enumitem}
\usepackage{soul}
\usepackage{mathtools,lipsum, nccmath,cuted,amsmath}
\usepackage[capitalise]{cleveref}
\usepackage{siunitx}
\usepackage{textcomp}

\usepackage{comment}
\usepackage{makecell}
\usepackage{algorithm}
\usepackage{gensymb}
\usepackage[normalem]{ulem}
\usepackage{subcaption}

\newtheorem*{example}{Example}
\let\oldexample\example
\renewcommand{\example}{\oldexample\normalfont}
\usepackage{listings}%
\definecolor{keywords}{HTML}{8A4A0B}

\definecolor{background}{HTML}{EEEEEE}

\definecolor{comments}{HTML}{868686}

\lstdefinelanguage{sdl}{
 morekeywords={scene,long,string,uniform,int,entity,'{','}'},
 keywordstyle=\color{keywords},
    basicstyle=\scriptsize\ttfamily,
 morecomment=[l]{//}, 
 morecomment=[s]{/*}{*/}, 
 morestring=[b]",
    basicstyle=\scriptsize\ttfamily,%
 commentstyle=\color{comments}\ttfamily,
numbers=right,
    numberstyle=\scriptsize,
    stepnumber=1,
    numbersep=8pt,
breaklines=true,
    frame=tb,
 tabsize=4}

\usepackage[font=footnotesize]{caption}
\usepackage{soul}
\usepackage{xcolor}
\usepackage{multirow}
\usepackage{soul}

\author{\IEEEauthorblockN{Charles Hartsell\textsuperscript{\textsection}, Shreyas Ramakrishna\textsuperscript{\textsection}, Abhishek Dubey, Daniel Stojcsics,  \\ Nagabhushan Mahadevan, and Gabor Karsai}
\IEEEauthorblockA{Institute for Software Integrated Systems, Vanderbilt University}\\
[-4.5ex]}

\DeclareAcronym{btd}{
  short = BTD,
  long  = Bow-Tie Diagram,
}
\DeclareAcronym{resonate}{
  short = ReSonAte,
  long  = Runtime Safety Evaluation in Autonomous Systems,
}
\DeclareAcronym{tfpg}{
  short = TFPG,
  long  = Timed Failure Propagation Graph,
}
\DeclareAcronym{gsn}{
  short = GSN,
  long  = Goal Structuring Notation,
}

\DeclareAcronym{cps}{
  short = CPS,
  long = Cyber Physical System,
}
\DeclareAcronym{dag}{
  short = DAG,
  long = Directed Acyclic Graph,
}
\DeclareAcronym{lec}{
  short = LEC,
  long = Learning Enabled Component,
}
\DeclareAcronym{les}{
  short = LES,
  long = Learning Enabled System,
}
\DeclareAcronym{aebs}{
  short = AEBS,
  long = Automatic Emergency Braking System,
}
\DeclareAcronym{sdl}{
  short = SDL,
  long = Scenario Description Language
}

\DeclareAcronym{srm}{
  short = SRM,
  long = Safety Risk Management
}

\DeclareAcronym{ood}{
  short = OOD,
  long = Out-Of-Distribution
}

\DeclareAcronym{rov}{
  short = ROV,
  long = Remote Operated Vehicle
}

\DeclareAcronym{uuv}{
  short = UUV,
  long = Unmanned Underwater Vehicle
}

\DeclareAcronym{av}{
  short = AV,
  long = Autonomous Vehicle
}

\DeclareAcronym{fact}{
  short = FACT,
  long = Fault Adaptive Control Technology
}
\DeclareAcronym{gps}{
  short = GPS,
  long = Global Positioning System
}
\DeclareAcronym{imu}{
  short = IMU,
  long = Inertial Measurement Unit
}
\DeclareAcronym{alarp}{
  short = ALARP,
  long = As Low As Reasonably Practicable
}

\begin{document}

\title{ReSonAte: A Runtime Risk Assessment Framework for Autonomous Systems}

\maketitle
\begingroup\renewcommand\thefootnote{\textsection}
\footnotetext{These Authors have equally contributed}
\endgroup

\pagestyle{plain}


\input{abstract.tex}

\input{paper}

\balance

\bibliographystyle{IEEEtran}
\bibliography{main.bib}
\end{document}

%% file: abstract.tex
\begin{abstract}
Autonomous \acp{cps} are often required to handle uncertainties and self-manage the system operation in response to problems and increasing risk in the operating paradigm. This risk may arise due to distribution shifts, environmental context, or failure of software or hardware components. Traditional techniques for risk assessment focus on design-time techniques such as hazard analysis, risk reduction, and assurance cases among others. However, these static, design-time techniques do not consider the dynamic contexts and failures the systems face at runtime. We hypothesize that this requires a dynamic assurance approach that computes the likelihood of unsafe conditions or system failures considering the safety requirements, assumptions made at design time, past failures in a given operating context, and the likelihood of system component failures. We introduce the ReSonAte dynamic risk estimation framework for autonomous systems. ReSonAte reasons over Bow-Tie Diagrams (BTDs) which capture information about hazard propagation paths and control strategies. Our innovation is the extension of the BTD formalism with attributes for modeling the conditional relationships with the state of the system and environment. We also describe a technique for estimating these conditional relationships and equations for estimating risk based on the state of the system and environment. To help with this process, we provide a scenario modeling procedure that can use the prior distributions of the scenes and threat conditions to generate the data required for estimating the conditional relationships. To improve scalability and reduce the amount of data required, this process considers each control strategy in isolation and composes several single-variate distributions into one complete multi-variate distribution for the control strategy in question. Lastly, we describe the effectiveness of our approach using two separate autonomous system simulations: CARLA and an unmanned underwater vehicle.

\end{abstract}


%% file: paper.tex
\section{introduction}
\label{sec:intro}

Autonomous Cyber Physical Systems (\acp{cps})\footnote{CPS with learning enabled components (LEC)} are expected to handle uncertainties and self-manage the system operation in response to problems and increase in risk to system safety. This risk may arise due to distribution shifts \cite{schwalbe2020survey}, environmental context or failure of software or hardware components \cite{koopman2016challenges}. \ac{srm} \cite{faa2000system} has been a well-known approach used to assess the system's operational risk. It involves design-time activities such as \textit{hazard analysis} for identifying the system's potential hazards,\textit{ risk assessment} to identify the risk associated with the identified hazards, and a system-level \textit{assurance case} \cite{bishop2000methodology} to argue the system's safety. However, the design-time hazard analysis and risk assessment information it uses is inadequate in highly dynamic situations at runtime. To better address the dynamic operating nature of \acp{cps}, dynamic assurance approaches such as runtime certification \cite{rushby2008runtime}, dynamic assurance cases \cite{denney2015dynamic}, and modular safety certificates \cite{schneider2013conditional} have been proposed. These approaches extend the assurance case to include system monitors whose values are used to update the reasoning strategy at runtime. A prominent approach for using the runtime information has been to design a discrete state space model for the system, identity the risk associated with each possible state transition action, then perform the action with the least risk \cite{wardzinski2008safety}.

The effectiveness of dynamic risk estimation has been demonstrated in the avionics \cite{kurd2009establishing} and medical \cite{leite2018dynamic} domains, but these techniques often encounter challenges with state space explosion which may limit their applicability to relatively low-complexity systems. Also, real-time \acp{cps} often have strict timing deadlines in the order of tens of milliseconds requiring any dynamic risk assessment technique to be computationally lightweight. Besides, the dynamic risk assessment technique should also take into consideration the uncertainty introduced by the \acp{lec} because of \ac{ood} data \cite{sundar2020out}. Assurance monitors \cite{cai2020real,sundar2020out} are a type of \ac{ood} detector often used to tackle the \ac{ood} data problem. The output of these monitors should be considered when computing the dynamic risk.

Recently, there is a growing interest in dynamic risk assessment of autonomous \ac{cps}. For example, the authors in \cite{katrakazas2019new} have used Dynamic Bayesian Networks to incorporate the broader effect of spatio-temporal risk gathered from road information on the system's operational risk. To perform dynamic risk assessment of a system with autonomous components, this paper introduces the \ac{resonate} framework. \ac{resonate} uses the design-time hazard analysis information to build \acp{btd} which describe potential hazards to the system and how common events may escalate to consequences due to those hazards. The risk posed by these hazards can change dynamically since the frequency of events and effectiveness of hazard controls may change based on the state of the system and environment. To account for these dynamic events at runtime, \ac{resonate} uses design-time \ac{btd} models along with information about the system's current state derived from system monitors (e.g. anomaly detectors, assurance monitors, etc.) and the operating environment (e.g. weather, traffic, etc.) to estimate dynamic hazard rates. The estimated hazard rate can be used for high-level decision making tasks at runtime, to support self-adaptation of \acp{cps}.


The specific contributions of this paper are the following. We present the \ac{resonate} framework and outline the dynamic risk estimation technique which involves design-time measurement of the conditional relationships between hazard rates and the state of the system and environment. These conditional relationships are then used at run-time along with state observations from multiple sources to dynamically estimate system risk.  Further, we describe a process that uses an extended \ac{btd} model for estimating the conditional relationships between the effectiveness of hazard control strategies and the state of the system and environment. To improve scalability and reduce the amount of data required, this process considers each control strategy in isolation and composes several single-variate distributions into one complete multi-variate distribution for the control strategy in question. A key contribution is our scenario specific language that enables data collection by specifying prior distributions over threat conditions and environmental conditions and initial conditions of the vehicle. 
We implement ReSonAte for an \ac{av} example in the CARLA simulator \cite{dosovitskiy2017carla} and through comprehensive simulations across 600 executions, we show that there is a strong correlation between our risk estimates and eventual vehicular collisions. The dynamic risk calculations on average take only 0.3 milliseconds at runtime in addition to the overhead introduced by the system monitors. Further, we exhibit ReSonAte's generalizability with preliminary results from an \ac{uuv} example.

\begin{figure*}[t]
\centering
 \includegraphics[width=0.9\textwidth]{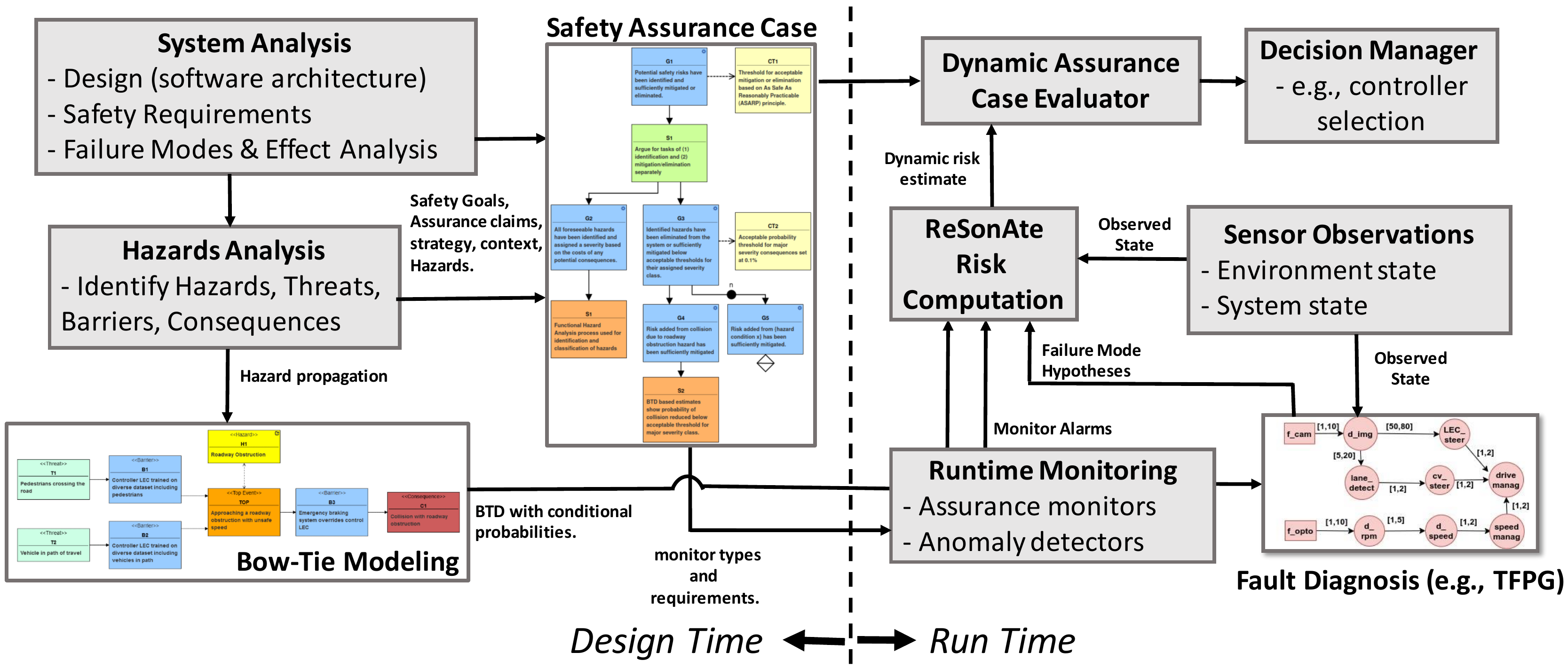}
 \caption{Overview of the steps involved in \ac{resonate}. Steps performed at design-time/run-time are shown on the left/right.}
 \label{fig:resonate-framework}
 \vspace{-0.1in}
\end{figure*}

\section{Related Work}
\label{sec:rw}

System health management \cite{mahadevan2011application,srivastava2011case} with model based reasoning have been popularly used for self-adaptation of traditional \acp{cps}. As discussed in \cite{steinbauer2013model}, a pre-requisite in these approaches has been that the system has knowledge about itself, its objectives, and its operating environments, including the ability to estimate when the adaption should occur. One way to do this is to estimate the operational risk to the system at runtime. Estimating the runtime risk is more difficult in autonomous systems with \acp{lec} due to the black box nature of the learning components and their susceptibility to distribution shifts and environmental changes. 

Our hypothesis is that we can use \acp{btd} derived from the system information and static assurance cases and use them to compose the anomaly and the threat likelihoods at runtime, including the likelihood for the failure of \acp{lec} \cite{sundar2020out,cai2020real,byun2020manifold,schwalbe2020survey}.   \acp{btd} are graphical models that provide a mechanism to learn the conditional relationships between threat events, hazards, and the success probability of barriers. \acp{btd} have been proactively used for qualitative risk assessment at design-time \cite{clothier2015structuring,williams2014building}, and has been recently combined with assurance arguments for operational safety assurance \cite{denney2017modeling,denney2019role}. 




Although \acp{btd} have been proactively used for risk assessment, there are several limitations in using them for quantitative risk computations, they are: (1) \textit{Static structure} - \acp{btd} have mostly been used as a graphical visualization tool for hazard analysis, and its static structure limits real-data updating which is required for dynamic risk estimation\cite{khakzad2012dynamic}, (2) \textit{Reliance on domain experts} - quantitative risk estimations mostly rely on domain experts to compute the probabilities for the \ac{btd} events such as threats, and barriers \cite{delvosalle2006aramis}, (3) \textit{Data uncertainty} - introduced because of data non-availability or insufficiency, and expert's limited knowledge makes it difficult to compute quality probability estimates \cite{ferdous2009handling}. 

Several approaches have been proposed to overcome these limitations and make \acp{btd} suitable for quantitative risk assessment. Bayesian techniques are one widely used approach that dynamically learns the \ac{btd} structure from design-time data, and updates the conditional probabilities for its events (e.g., threats, barriers) \cite{bayesian-btd,badreddine2013bayesian}. Further, \cite{teimourikia2017run} transforms a \ac{btd} into a Bayesian Network where each node of the \ac{btd} is modelled as a Bayesian Node. The other approach uses techniques such as fuzzy set and evidence theory to address the data uncertainty problems \cite{ferdous2009handling,zhang2018selecting}. In this approach, the expert's knowledge is translated into numerical quantities that are used in the \acp{btd}. \cite{vileiniskis2017quantitative} extends \acp{btd} with petri-nets models and monte-carlo simulations to generate data. 


Although prior approaches have made \acp{btd} suitable for quantitative risk estimation, the design-time hazard information used for risk estimation is inadequate for \acp{cps}. The main focus of this work is to dynamically estimate risk by fusing design-time information captured in the \acp{btd} with run-time information about the system and the environment. Additionally, we concentrate on generating large-scale simulation data for conditional probability estimation of the \ac{btd} events.

\input{induvidual-sections/framework}

\input{induvidual-sections/evaluation}

\section{Conclusion and Future Work}
\label{sec:conclusion}

\ac{resonate} captures design-time information about system hazard propagation, control strategies, and potential consequences using \acp{btd}, then uses the information contained in these models to calculate risk at run-time. The frequency of threat events and the effectiveness of hazard control strategies influence the estimated risk and may be conditionally dependent on the state of the system and operating environment. A technique for measuring these conditional relationships in simulation using a custom \ac{sdl} was demonstrated. \ac{resonate} was then applied to an \ac{av} example in the CARLA simulator to dynamically assess the risk of collision, and a strong correlation was found between the estimated likelihood of a collision and the observed collisions. Additionally, \ac{resonate}'s risk calculations require minimal computational resources making it suitable for resource-constrained and real-time \acp{cps}.

Future extensions and applications for \ac{resonate} include: (1) dynamic estimation of event severity in addition to event likelihood, (2) inclusion of state uncertainty into risk calculations to produce confidence bounds on risk estimates, (3) forecasting future risk based on expected changes to system or environment, (4) use of estimated risk for higher-level decision making such as controller switching or enactment of contingency plans, and (5) continuous improvement of conditional probability estimates at run-time from operational data.

\textbf{Acknowledgement}: This work was supported by the DARPA Assured Autonomy project and Air Force Research Laboratory. 



%% file: induvidual-sections/framework.tex
\section{ReSonAte Framework}
\label{sec:framework}
The goal of the \ac{resonate} framework is the dynamic estimation of \textit{risk} based on runtime observations about the state of the system and environment. We define risk as a product of the likelihood and severity of undesirable events, or \textit{consequences}. \cref{fig:resonate-framework} outlines the ReSonAte workflow which is divided into design-time and run-time steps. The design-time steps include System Analysis, Hazard Analysis, Assurance Case Construction, and \ac{btd} Modeling. Sections \ref{sec:sa} through \ref{subsec:bowtie_formalization} provide background information about each of these well-studied techniques. However, our \ac{btd} formalism described in \cref{subsec:bowtie_formalization} is distinct from existing formalisms with additional model attributes and restrictions for the \ac{resonate} framework. Sections \ref{subsec:risk_calculation} through \ref{sec:framework/dynamic_assurance} introduce the risk calculation equations, the conditional probability estimation process, and the dynamic assurance case evaluation method of the \ac{resonate} framework respectively.


\subsection{Background}

\subsubsection{System Analysis}
\label{sec:sa}
System analysis involves the design-time analysis of the system operation, system faults, the available runtime monitors, and the operating environment such as weather (e.g. rain, snow, fog, etc.), traffic, etc. Thereafter, we perform failure mode analysis to identify the possible component faults and their potential impacts on the safety of the system. Fault propagation paths can be described with the use of an appropriate fault modeling language (e.g. \ac{tfpg} \cite{misra1994senor}), and run-time monitors for fault identification can be designed as appropriate. Monitors to detect faults in traditional components have been designed in prior work \cite{mahadevan2012architecting}, but monitors for \acp{lec} are often more complex (e.g. assurance monitors \cite{sundar2020out}). Alarms raised by these monitors at run-time can then be fed into an appropriate fault diagnosis engine (e.g. the \ac{tfpg} reasoning engine) to isolate particular fault modes that may be present in the system.

\subsubsection{Hazard Analysis}
\label{subsec:hazard_analysis}
Hazard analysis involves the identification of events that may lead to hazardous conditions, implementation of barriers to prevent loss of control over hazard conditions or recovery control after a loss, and estimation of the risk posed by the identified hazards. Guidelines for performing hazard analysis are available in a variety of domains including the ISO 26262 standard \cite{iso26262} for the automotive domain and the FAA System Safety Handbook \cite{faahandbook} for the aerospace domain.

\subsubsection{Assurance Case Construction}
An assurance case \cite{bishop2000methodology} is a structured argument supported by a body of evidence which shows that system goals will be met in a defined operating environment and is often documented using the \ac{gsn} \cite{kelly1999arguing}. Multiple authors have noted the importance of having "hazard analysis and risk reduction arguments" within an assurance case \cite{haddon2009nimrod},\cite{leveson2011use}. In this work, we only concentrate on the hazard analysis argument of the assurance case, and earlier works \cite{matsuno2012toward,denney2015dynamic} can be referred for a more comprehensive dynamic assurance case.

\begin{figure*}[t]
    \centering
    \includegraphics[width=\textwidth]{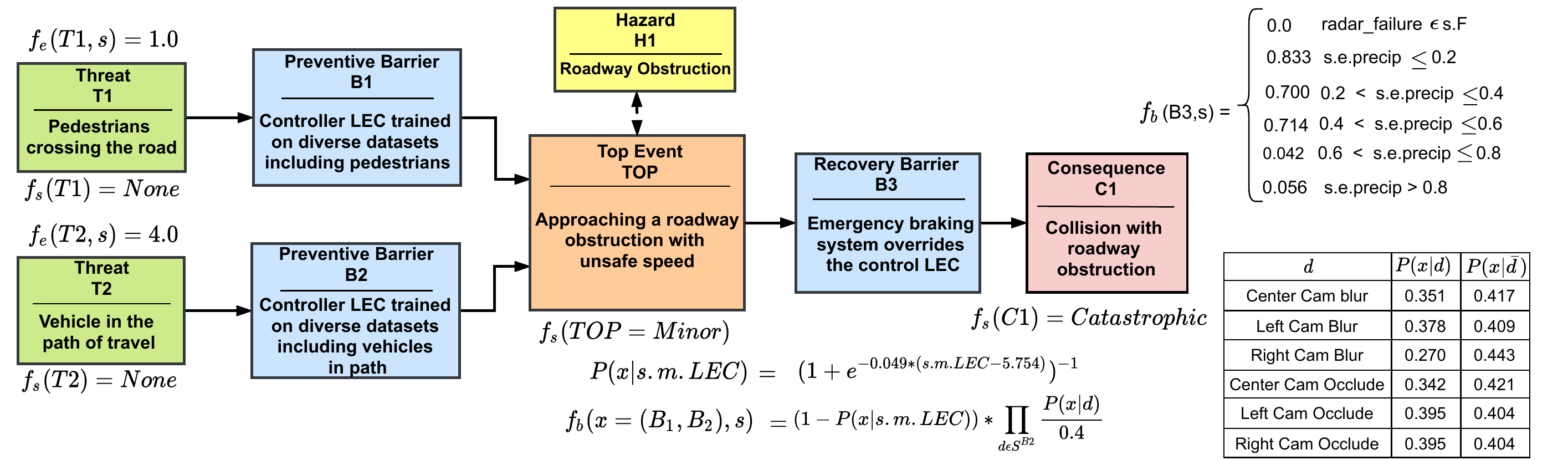}
    \caption{Example Bow-Tie Diagram for autonomous vehicle example "Roadway Obstruction" hazard created using the ALC Toolchain. Each block includes a brief description and the type of each node is denoted at the top of the block. The equations depicted are described in \cref{subsec:bowtie_formalization}.}
    \label{fig:bowtie-example}
    \vspace{-0.1in}
\end{figure*}

\subsection{Our Contributions: Bow-Tie Formalization \& Extensions}
\label{subsec:bowtie_formalization}


\acl{btd}s are used in \ac{resonate} as the means of describing hazard propagation paths and the control strategies used to prevent that propagation. \acp{btd} are intended for describing linear propagation and do not have concepts for capturing non-linear causality or complex interactions between events. However, these linear models are sufficient for many hazards commonly encountered by \acp{cps} such as those demonstrated for the example systems described in \cref{sec:evaluation}. To perform dynamic risk estimation, the hazard models must also contain information about the expected rate of threat events and the success probability of barriers, both of which may be conditional on the current conditions. In this section, we present a \ac{btd} formalization which includes this conditional information not captured in existing \ac{btd} interpretations. Under our formalism, each node has an associated function that is conditional on the state of the system and environment. Before constructing a \ac{btd}, we assume the following sets have been identified: all potential hazard classes $H$, all events $E$ relevant for hazard analysis, barriers $B$ which may either \textit{prevent} a loss of control or \textit{recover} after such a loss has occurred, possible system failure modes $FM$, and event severity classes $SV$. We also assume a function $f_a$ which maps each severity class to the maximum acceptable rate of occurrence threshold $f_a : SV \rightarrow \mathbb{R}$ has been defined.

For risk calculation, we define the state of the system and the environment as $S = (F, e, m)$ where: 


\begin{itemize}
    \item $F$ is a set of failure modes currently present in the system. $F \subseteq FM$
    \item $e$ is an n-tuple describing the current environmental conditions where $n$ is the number of environmental parameters that are relevant for risk calculations and each parameter may be continuous or discrete valued.
    \item $m$ is a k-tuple containing the outputs of runtime monitors in the system where $k$ is the number of such monitors and each monitor may be continuous or discrete valued. 
\end{itemize}

When modeling the environment, the system developer should choose an appropriate level of abstraction based on which environmental parameters are relevant for risk calculation. More environmental parameters may increase the fidelity of the model but also increases the required effort when determining conditional probabilities for events and barriers. It is not required to consider every environmental parameter for all events and barriers. Instead, each event or barrier may be conditional upon a smaller subset of environmental parameters that have the largest impact on that particular event or barrier. Formally, we define a \acl{btd} as a tuple $(N, C, f_t, h, f_d, f_b, f_e, f_s)$ where:

\begin{itemize}
    \item $N$ is a set of nodes where each node represents either an event $e \in E$ or a barrier $b \in B$.
    \item $C$ represents a set of directed connections such that $C \subseteq N \times N$. For a given connection $c \in C$, $src(c)$ and $dst(c)$ represent the source and destination of $c$ respectively.
    \item The tuple $(N, C)$ is a \ac{dag} representing the temporal ordering of events as well as the barriers which may break this ordering and prevent further propagation of events.
    \item The function $f_t$ gives the type of each node. $f_t : N \rightarrow \{event, barrier\}$. Using this function, we let $N_e = \{n \in N\ |\ f_t(n) = event\}$ and $N_b = \{n \in N\ |\ f_t(n) = barrier\}$. This gives the following properties: $N_e \subseteq E$, $N_b \subseteq B$, $N = N_e \cup N_b$, and $N_e \cap N_b = \emptyset$.
    \item $h \in H$ is the hazard class associated with the \ac{btd}.
    \item The function $f_d$ gives a textual description of each node. $f_d : N \rightarrow string$
    \item $f_b$ is a probability function conditional on the state of the system and environment defined for each barrier node. This function represents the probability that the barrier will successfully prevent further event propagation. $f_b: N_b \times S \rightarrow [0, 1]$. 
    \item $f_e$ is a function conditional on the state of the system and environment defined for each event node. It gives the expected frequency of the event over a fixed period. The values of this function must be specified for \textit{threat} events (i.e. root events with no preceding input events), but can be calculated for subsequent events in the diagram as discussed in Section \ref{subsec:risk_calculation}. $f_e: N_e \times S \rightarrow [0, \inf)$. 
    \item A function $f_s$ which maps each event to the appropriate severity class. $f_s : N_E \rightarrow SV$
\end{itemize}
    
\acp{btd} are often constructed in a chained manner where a consequence event in one section of the \ac{btd} may serve as a threat or top event in a subsequent section of the same \ac{btd}. However, such a chained \ac{btd} can be broken into multiple, single-scope \acp{btd} where each event has one unique type (i.e. threat, top event, or consequence). As a simplification, \ac{resonate} operates only on these single-scope \acp{btd} which satisfy the additional restrictions listed below. Note that the symbol $\Rightarrow$ is used to denote precedence in the graph. That is, given two nodes $a, b \in N$, then $a \Rightarrow b$ states that $a$ precedes $b$ in the \ac{btd}, but this does not necessitate a direct edge such that $a \rightarrow b$. Instead, there may be any number of intermediate nodes $c_i \in N$ such that $a \rightarrow c_1 \rightarrow c_2 \rightarrow ... \rightarrow c_n \rightarrow b$.

\begin{itemize}
    \item Exactly one event must be designated as the \textit{top event}, denoted $e_{top}$.
    \item There must be at least one \textit{threat}. i.e. $\exists\ t \in N_e\ |\ t \Rightarrow e_{top} \land \nexists\ n \in N\ |\ n \Rightarrow t$. We denote the set of all threat events as $N_t$.
    \item There must be at least one \textit{consequence}. i.e. $\exists\ c \in N_e\ |\ e_{top} \Rightarrow c \land \nexists\ n \in N\ |\ c \Rightarrow n$. We denote the set of all consequence events as $N_c$.
    \item All events must be a threat, a top event, or a consequence. i.e. No intermediate events.
    \item All \textit{barriers} must lie between a threat and the top event, or between the top event and a consequence. i.e. $\forall\ b \in N_b$ either $t \Rightarrow b \Rightarrow e_{top}$ or $e_{top} \Rightarrow b \Rightarrow c$
    \item No branching or joining of the graph is allowed, except for at the top event.
\end{itemize}

For the example \ac{btd} shown in \cref{fig:bowtie-example}, the type of each event is denoted on the top of the block and we can define the sets $N_e$ = $\{T1, T2, TOP, C1\}$ and $N_b$ = $\{B1, B2, B3\}$. For each event $e \in N_e$, the associated severity class is given by the function $f_s(e)$ which is shown under each event block. Each of the threats, $T1$ and $T2$, have been assigned to the "None" severity class because these events are considered to be common occurrences that do not result in any safety violation by themselves. The top event $TOP$ has been assigned to the "Minor" severity class since this event can still be mitigated before a safety violation occurs but mitigation requires the \ac{aebs} to override the primary LEC controller. Finally, the consequence $C1$ has been assigned to the "Catastrophic" severity class since this event is a safety violation and may result in significant damage to the system or environment. 

The conditional functions $f_e$ and $f_b$ are shown near their respective nodes in \cref{fig:bowtie-example}. The functions $f_e(T1, \textbf{s})$ and $f_e(T2, \textbf{s})$ give the expected frequency of threats $T1$ and $T2$ in units of expected number of occurrences per minute, and their values were measured for our simulator configuration described in \cref{sec:evaluation}. The probability function for barriers $B1$ and $B2$, $f_b$(x = (B1,B2), \textbf{s}), shows how these barriers are dependent on both the continuous-valued output from the assurance monitor and on the binary state of other monitors. A sigmoid function, $P(x | \textbf{s}.m.LEC)$, is used to capture the conditional relationship with the assurance monitor output. Finally, $f_b(B3, \textbf{s})$ shows how barrier $B3$ is less likely to succeed as the precipitation increases and will not function in the case of a radar failure. \cref{subsec:risk_calculation} explains the generic process for conditional probability estimation and \cref{sec:evaluation/conditional_relationships} provides more detail on the functions in this \ac{btd}.

\subsection{Run-time Risk Computation}
\label{subsec:risk_calculation}
Each \ac{btd} includes functions describing the conditional frequency for all \textit{threat} events and the conditional probability of success for all \textit{barrier} nodes. However, the likelihood of each \textit{consequence} is necessary to estimate the overall level of risk for the system. These probabilities can be calculated by the propagation of the initial threat rates through the \ac{btd}. When a particular event occurs, barrier nodes reduce the probability that the event will continue to propagate through the \ac{btd} based on the following equation:

\begin{gather}
R(e_2 | s) = R(e_1 | s) P(\overline{b_1} \land \overline{b_2} \land ... \land \overline{b_n} | s) \nonumber \\
assume\ a \perp b\ \forall\ a, b \in \{b_1, b_2, ..., b_n\}\ |\ a \neq b \nonumber \\
\label{eqn:conditional_rate}
R(e_2 | s) = R(e_1 | s) [\Pi_{i=1}^n P(\overline{b_i} | s)]
\end{gather}

where $e_1, e_2 \in N_e$ and $b_i \in N_b$ such that $e_1 \rightarrow b_1 \rightarrow b_2 \rightarrow ... \rightarrow b_n \rightarrow e_2$. $R(e_i | s)$ represents the frequency of event $e_i$ given state $s$. \cref{eqn:conditional_rate} makes the assumption that no barriers share any common failure modes and that the effectiveness of each barrier is independent from the outcome of other barriers. Similarly, $P(\overline{b_i} | s)$ represents the probability that barrier $b_i$ will fail to prevent event propagation given state $s$, i.e. $P(\overline{b_i} | s)$ = $1 - P(b_i | s)$. Letting $S$ represent the set of all states we are concerned with, then we can calculate the overall frequency of $e_2$ as $R(e_2) = \Sigma_{s \in S}[ R(e_2 | s) P(s) ]$ where $P(s)$ is the probability of each particular state $s \in S$. As discussed in the \ac{btd} formalization in Section \ref{subsec:bowtie_formalization}, no joining or splitting of paths is allowed in a \ac{btd} with the exception of the top event $e_{top}$. We treat the top event as a summation operation for all incoming edges, i.e. any threat event may independently cause a top event if the associated barriers are unsuccessful. All outgoing edges from the top event are treated as independent causal chains, i.e. any potential consequence may occur from a top event, independent of other consequences in the \ac{btd}. The probability for the top event can be calculated as:



\begin{gather}
R(t^{top}) = \Sigma_{s \in S}[ R(t | s) P(s) [\Pi_{i=1}^n P(\overline{b_i} | s)] ] \nonumber \\
\label{eqn:top_event_rate}
R(e_{top}) = \Sigma_{t \in N_t} R(t^{top})
\end{gather}



where $R(e_{top})$ is the frequency for the top event and $P(t^{top})$ represents the contribution of each threat $t$ to this rate after passing through any intermediate prevention barriers $b_i \in N_b$ such that $t_i \rightarrow b_1 \rightarrow b_2 \rightarrow ... \rightarrow b_n \rightarrow e_{top}$. Finally, the probability of each consequence can be calculated with:

\begin{equation}
\label{eqn:consequence_rate}
    R(c_i) = R(e_{top}) \Sigma_{s \in S}[P(s) [\Pi_{i=1}^n P(\overline{b_i} | s)] ]
\end{equation}

where $R(c_i)$ is the frequency of consequence $c_i \in N_c$ after passing through any recovery barriers $b_i \in N_b$ such that $e_{top} \rightarrow b_1 \rightarrow b_2 \rightarrow ... \rightarrow b_n \rightarrow c_i$.

If the state is not known uniquely, then each potential state $\textbf{s} \in S$ must be enumerated and a probability function $P(\textbf{s})$ must be assigned such that $\Sigma_{\textbf{s} \in S} P(\textbf{s}) = 1$. At design-time when only the expected distributions of the state variables are known, this state probability function is typically calculated as a product of the probability mass functions (or probability density functions for continuous variables) of each individual state variable. Recall that for \ac{resonate} the state $S$ is restricted to only those variables which have a conditional impact on the functions contained in the \acp{btd} and thus not all state variables describing the system must be considered in this calculation. When the system is deployed at run-time, observations about the current state can be used to refine the set of prior probabilities $P(s)$ and dynamically calculate current risk values. The equations outlined here use a discrete treatment of probability, but continuous distributions can be used as well where summation operations are replaced by an appropriate integration. If the state can be identified uniquely to a particular state $\textbf{s}_0 \in S$, then we can assign $P(\textbf{s}_0) = 1$ and simplify the risk equations. For example, we could calculate the rate of occurrence for events $TOP$ and $C1$ which are part of the \ac{btd} $B$ shown in \cref{fig:bowtie-example} using the following equations:





\label{eqn:top_event_rate_example}
\begin{align*}
    R(TOP | s_0) &= B.f_e(T1, s_0) * (1 - B.f_b(B1, s_0))\\
    & + B.f_e(T2, s_0) * (1 - B.f_b(B2, s_0))
\end{align*}

\begin{equation}
\label{eqn:consequence_rate_example}
    R(C1 | s_0) = R(TOP | s_0) * [1 - B.f_b(B3, s_0)]
\end{equation}

\subsection{Estimating Conditional Relationships}
\label{subsec:conditional_estimation}

\subsubsection{Conditional Relationships}
In \ac{resonate}, both the rate of occurrence of events and the effectiveness of barriers may be conditionally dependent on the state including system failure modes, runtime monitor values, and environmental conditions. For each of these categories, it is necessary to identify the factors which should be examined for their impact on this conditional relationship. A failure analysis should be performed using an appropriate failure modeling language as discussed in \cref{sec:sa}. Similarly, the environmental parameters relevant to the system must be identified and the operating environment defined in terms of bounds and expected distributions of these parameters. We assume the environmental conditions are known uniquely and provided to \ac{resonate}. Monitor values that may impact the conditional relationships should also be identified.

For some nodes in a \ac{btd} it may be possible to analytically derive the appropriate conditional relationship with each state variable, but often this relationship must be inferred from data. In this section, we consider a generic threat to the top event chain with a single barrier described as $t \rightarrow b \rightarrow e_{top}$. The contribution to the rate of the top event $e_{top}$ from this singular threat $t$ with a single barrier $b$ is shown in \cref{eqn:b_event_chain}. Normally, multiple threats can lead to the top event and the contribution of all threats must be considered as described in \cref{eqn:top_event_rate}. However, if all other threat conditions can be eliminated, then the top event may only occur as a result of threat $t$. In this case, the probability of success for barrier $b$ can be calculated as a function of the ratio of frequencies between the top event and threat $t$ shown in \cref{eqn:b_probability}. This approach can also be used for threats with multiple associated barriers by considering each barrier in isolation since individual barriers are assumed to be independent in \cref{eqn:conditional_rate}. 

\begin{gather}
\label{eqn:b_event_chain}
R(e_{top}^t | \textbf{s}) = R(t | \textbf{s}) * (1 - P(b | \textbf{s})) \\
R(t_j | \textbf{s}) = 0\ \forall\ t_j \in N_t\ |\ t_j \neq t \rightarrow  R(e_{top} | \textbf{s}) = R(e_{top}^t | \textbf{s}) \nonumber \\
\label{eqn:b_probability}
P(b | \textbf{s}) = 1 - [R(e_{top} | \textbf{s})/R(t | \textbf{s})]
\end{gather}

We isolate individual threats in simulation using a custom \ac{sdl}, described in \cref{sec:sdl}, to generate scenarios where only the one threat of interest is allowed to occur and all other possible threats are eliminated. Each time the threat $t$ is encountered, the top event $e_{top}$ may either occur or not occur. If the top event does not occur, then the associated barrier $b$ was successful - i.e. prevented hazard propagation along this path. Otherwise, the barrier was unsuccessful. Since the occurrence of a threat is a discrete event that results in a boolean outcome (i.e. top event does/does not occur), the barrier effectiveness can be modeled as a conditional Binomial distribution where the probability of barrier success is dependent on the ratio of top event frequency to threat frequency as shown in \cref{eqn:b_probability}. 


For each barrier $b$ in the \ac{btd}, any state variables which are likely to impact the conditional frequency or probability of that node should be identified, and we denote this reduced set of state variables as $S^b$. For discrete state variables (e.g., presence or absence of a particular fault condition, urban/rural/suburban environment, etc.), this ratio can be estimated using Laplace's rule of succession \cite{zabell1989rule} as shown in \cref{eqn:rule_of_succession} where $n_{s_i = a}$ is the number of scenes where the state variable $s_i$ had the desired value $a$ and $k_{s_i = a}$ is the number of such scenes where the top event also occurred. This equation can be applied for each of the possible values of the state variable $s_i$ to estimate the discrete probability distribution $P(b | s_i)$, and the process can then be repeated for each relevant state variable $s_i \in S^c$. \cref{eqn:multivariate_dist} can be used to fuse the estimated distributions of each individual state variable into one multivariate distribution $P(b | \textbf{s})$. Similar to Naive Bayes classifiers, this equation assumes each of the state variables $s_i$ are mutually independent conditional on the success of barrier $b$. If a stronger assumption is used that the state variables in $S^{b}$ are mutually independent, then the term $\Pi_{j = 0}^m [P(s_j)] P(s_0 s_1 ... s_m)^{-1}$ reduces to 1 as was the case for all of the probabilities estimated in our examples. For each continuous state variable $s_j$ (e.g. output of an assurance monitor), maximum likelihood estimation was used in place of \cref{eqn:rule_of_succession} to estimate $P(b | s_j)$. Similarly, \cref{eqn:multivariate_dist} can be revised for continuous values by replacing the probability mass functions $P(s_j)$ with probability density functions $p(s_j)$.

\begin{gather}
\label{eqn:rule_of_succession}
P(b | s_i = a) = 1 - \frac{k_{s_i = a} + 1}{n_{s_i = a} + 2} \\
P(b | \textbf{s}) = \frac{P(s_0, s_1, ..., s_m | b)P(b)}{P(s_0, s_1, ..., s_m)} \nonumber \\
assume\ s_j \perp s_k\ |\ b\ \forall\ s_j, s_k \in S^{b}\ |\ s_j \neq s_k \nonumber \\
P(b | s_0, s_1, ..., s_m) = \frac{P(b) \Pi_{j = 0}^m P(s_j | b)}{P(s_0, s_1, ..., s_m)} \nonumber \\
\label{eqn:multivariate_dist}
P(b | s_0, s_1, ..., s_m) = \frac{\Pi_{j = 0}^m P(b | s_j) P(s_j)}{P(b)^{m-1} P(s_0 s_1 ... s_m)}
\end{gather}






While the conditional probability estimation process is outlined here for prevention barriers, it may be modified for recovery barriers by replacing occurrences of any threats \textit{t} with the top event $e_{top}$ and replacing occurrences of the top event with each consequence of interest.

\subsubsection{Scenario Description Language}
\label{sec:sdl}
Description and generation of scenarios that cover the full range of expected operating environments is an important aspect for the design of \ac{cps}. Several domain-specific \acp{sdl} such as Scenic \cite{fremont2019scenic} and MSDL \cite{msdl} with probabilistic scene generation capabilities are available. While these languages have powerful scene generation capabilities, they are targeted specifically at the automotive domain. We have developed a simplified \ac{sdl} using the textX \cite{dejanovic2017textx} meta language to generate varied scenes for multiple domains including our \ac{av} and \ac{uuv} systems. 



\input{scene-example}

A fragment of the scene description for the CARLA \ac{av} example is shown in \cref{fig:example-scene}. A scene S = \{$e_1$,$e_2$,...,$e_i$\} is described as a collection of entities (or set points), with each entity representing information either about the ego vehicle (e.g., type, route, etc.), or the operating environment (e.g., weather, obstacle). Further, each of these entities has parameters whose value can be sampled using techniques such as Markov chain Monte Carlo to generate different scenes in the simulation space. The larger the number of sampling, the wider is the simulation space coverage. For the \ac{av} example, our scene was defined as S = \{town\_description, weather\_description, av\_route\}. While the parameters of town\_description and av\_route remained fixed, the parameters of weather\_description such as cloud, precipitation, and precipitation deposits were randomly sampled to take a value in [0,100]. We generated 46 different CARLA scenes by randomly sampling the weather parameters, a few of which were used for estimating the conditional relationships. 

Currently we perform unbiased sampling to generate each scenario, then use the resulting unbiased dataset for the probability calculations described in \cref{subsec:conditional_estimation}. This approach proved sufficient for the example systems described in \cref{sec:evaluation}. However, these systems are prototypes where consequences occur relatively often. For more refined production systems, consequences do not typically occur under nominal operation but instead are often the result of rare combinations of adverse operating conditions and/or system failure modes. This is an example of the long tails problem where the probability of observing this undesired system behavior is low if unbiased random sampling is used. In future work, guided sampling of the state space (e.g. \cite{karunakaran2020efficient}) can be used to better observe these rare events and perform conditional probability estimation with the resulting biased dataset.



\subsection{Dynamic Assurance Case Evaluation}
\label{sec:framework/dynamic_assurance}
\acl{srm} \cite{faa2000system} is a common technique used in the system safety assurance process which involves identification of potential hazards, analysis of the risks posed by those hazards, and reduction of these risks to acceptable levels. The amount of risk remaining after risk control strategies have been implemented is known as the \textit{residual risk}. An appropriate assurance argument pattern, such as the \ac{alarp} pattern outlined by Kelly \cite{kelly1999arguing}, is often used to document the means used for risk reduction and show that the estimated levels of residual risk are within tolerable bounds. With traditional \ac{srm} techniques, residual risk estimates are static results of design-time analysis techniques. However, using the risk estimated by \ac{resonate} the residual risk for each hazard can be updated dynamically at run-time and the associated goal in the assurance argument can be invalidated if the risk exceeds a predefined threshold. When this risk threshold is violated, contingency plans can be enacted to place the system in a safe state. For example, stopping the AV or surfacing the UUV are simple contingency actions used for the example systems described in \cref{sec:evaluation}. The risk scores produced by \ac{resonate} may also be used in assurance case adaptation techniques such as Dynamic Safety Cases \cite{denney2015dynamic} or ENTRUST \cite{calinescu2017engineering}.

%% file: scene-example.tex
\begin{figure}[!t]
\setlength{\abovecaptionskip}{-3pt}
\begin{lstlisting}[language=sdl,numbers=none]
scene sample {
    type string
    type int
    entity town_description{
        id:string
        map:string  }
    entity weather_description{
        cloudiness: uniform
        precipitation: uniform
        precipitation_deposits: uniform  }
    entity uniform{
        low: int
        high: int  }
	}
 
\end{lstlisting}
\caption{This listing shows a fragment of a CARLA scene description that was generated using our \ac{sdl} written in textX meta language.  
}
\label{fig:example-scene}
 \vspace{-0.2in}
\end{figure}

%% file: induvidual-sections/evaluation.tex
\section{Evaluation}
\label{sec:evaluation}
We evaluate \ac{resonate} using an \ac{av} example in the CARLA simulator \cite{dosovitskiy2017carla} and show its generalizability with preliminary results from an \ac{uuv} example \cite{robotics2016bluerov2}. The experiments\footnote{\label{github}source code to replicate the CARLA \ac{av} experiments can be found at: \url{https://github.com/scope-lab-vu/Resonate}} in this section were performed on a desktop with AMD Ryzen Threadripper 16-Core Processor, 4 NVIDIA Titan Xp GPU's and 128 GiB memory.

\begin{figure}[t]
\centering
    \begin{subfigure}{.49\columnwidth}
      \centering
      \includegraphics[width=\linewidth]{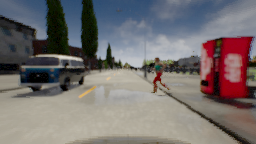}
      \caption{Screenshot from CARLA simulator.}
      \label{fig:carla_screenshot}
    \end{subfigure}
    \begin{subfigure}{.42\columnwidth}
      \centering
      \includegraphics[width=\linewidth]{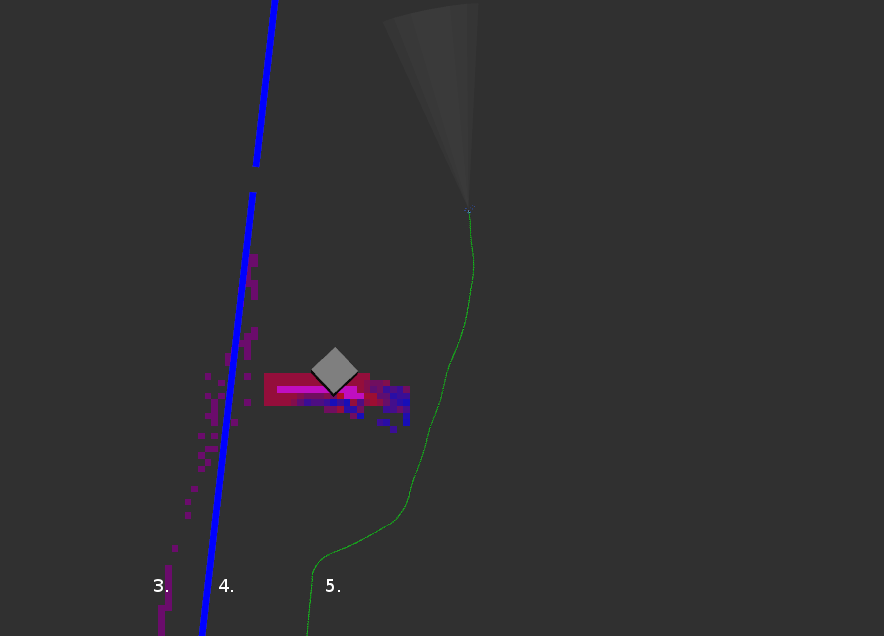}
      \caption{Top-down view of BlueROV2.}
      \label{fig:bluerov_screenshot}
    \end{subfigure}
 \caption{Screenshots from each autonomous system simulation. \cref{fig:carla_screenshot} shows an image from the forward-looking camera of the \ac{av} as it navigates through the city. \cref{fig:bluerov_screenshot} shows a top-down view of the \ac{uuv} where the vehicle (trajectory shown as a green line) is inspecting a pipeline (thick blue line) until an obstacle (grey box) is detected by the forward-looking sonar and the vehicle performs an avoidance maneuver.}
 \label{fig:simulator_screenshots}
\vspace{-0.1in}
\end{figure}

\subsection{Autonomous Ground Vehicle}

\subsubsection{System Overview}
\label{sec:sys}

\begin{figure}[t]
 \centering
 \includegraphics[width=0.9\columnwidth]{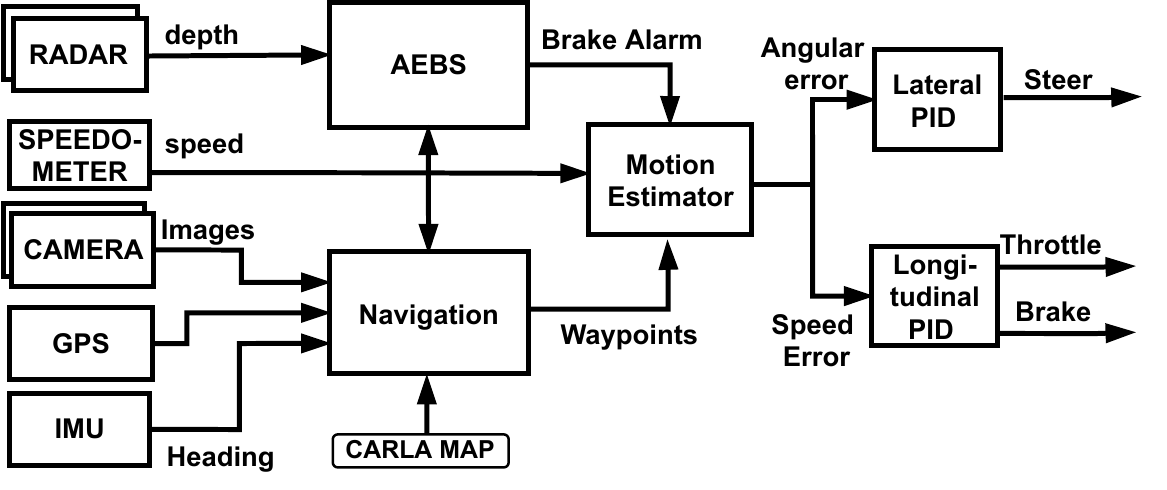}
 \caption{A block diagram of our \ac{av} example in CARLA simulation.}
 
 \label{fig:sys-model}
  \vspace{-0.15in}
\end{figure}

Our first example system is an autonomous car which must safely navigate through an urban environment while avoiding collisions with pedestrians and other vehicles in a variety of environmental and component failure conditions. The architecture of our \ac{av}, shown in \cref{fig:sys-model}, relies on a total of 9 sensors including two forward-looking radars, three forward-looking cameras, a \ac{gps} receiver, an \ac{imu}, and a speedometer. The "Navigation" \ac{lec}, adapted from previous work \cite{chen2020learning}, produces waypoints for the desired position and velocity of the vehicle at a sub-meter granularity using a neural network for image processing along with higher-level information about the desired route provided by a map. These waypoints are passed to the "Motion Estimator" which computes throttle and steering angle error between the current and desired waypoints. The Motion Estimator also serves as a supervisory controller which will override the primary Navigation component if an alarm is sent by the \ac{aebs} safeguard component. This \ac{aebs} component will raise an alarm when the vehicle safe stopping distance, estimated based on the current vehicle speed, exceeds the distance returned by the radars indicating that the \ac{av} is approaching an object at an unsafe speed. Finally, the output of the Motion Estimator is sent to two PID controllers which generate appropriate steering, throttle, and brake control signals.



\subsubsection{System Analysis}
\label{subsec:system_setup}
We start with the analysis of the \ac{av} system and its operating environment (e.g., weather, traffic, etc). As the \ac{av} primarily uses a perception \ac{lec}, parameters such as weather conditions (e.g., cloud spread, precipitation level, and precipitation deposit level), high brightness, and camera related faults such as blur and occlusion influenced the control actions generated by the LEC controller. To detect the camera faults and adverse operating conditions, we have designed several monitors. Each of the three cameras is equipped with OpenCV based blur and occlusion detectors to detect image distortions. The blur detector uses the variance of the laplacian \cite{pech2000diatom} to quantify the level of blur in the image where a high variance indicates that the image is not blurred, while a low value ($<$30) indicates the image is blurred. The occlusion detector is designed to detect continuous black pixels in the images, with the hypothesis that an occluded image will have a higher percentage of connected black pixels. In this work, a large value ($>$15\%) of connected black pixels indicated an occlusion. The blur detector has an F1-score of 99\% and the blur detector has an F1-score of 97\% in detecting the respective anomalies.


Additionally, we leverage our previous work \cite{sundar2020out} to design a reconstruction based $\beta$-VAE assurance monitor for identifying changes in the operating scenes such as high brightness. The $\beta$-VAE network has four convolutional layers $32/64/128/256$ with (5x5) filters and (2x2) max-pooling followed by four fully connected layers with $2048$, $1000$, and $250$ neurons. A symmetric deconvolutional decoder structure is used as a decoder, and the network uses hyperparameters of $\beta$=1.2 and a latent space of size=100. This network is trained for 150 epochs on 6000 images from CARLA scenes of both clear and rainy scenarios. The reconstruction mean square error of the $\beta$-VAE is used with Inductive Conformal Prediction \cite{shafer2008tutorial} and power martingale \cite{fedorova2012plug} to compute a martingale value. The assurance monitor has an F1 score of 98\% in detecting operating scenes with high brightness. Further, \ac{tfpg} models for the identified camera faults were constructed, but the \ac{tfpg} reasoning engine was not used since the available monitors were sufficient to uniquely isolate fault conditions without any additional diagnostic procedures.

\subsubsection{Hazard Analysis and \ac{btd} Modeling}
\label{sec:hazard}
For our \ac{av} example, we consider a single hazard of a potential collision with roadway obstructions. The potential threats for this hazard were identified to be pedestrians crossing the road (T1), and other vehicles in the \ac{av}'s path of travel (T2). The top event was defined as a condition where the \ac{av} is approaching a roadway obstruction with an unsafe speed. The primary \ac{lec} is trained to safely navigate in the presence of either of these threats and served as the first hazard control strategy for both threat conditions, denoted by prevention barriers B1 and B2. Finally, the \ac{aebs} served as a secondary hazard control strategy denoted by the recovery barrier B3. The \ac{btd} shown in \cref{fig:bowtie-example} was constructed based on these identified events and control strategies. For full-scale systems, the \ac{btd} construction process will usually result in the identification of a large number of events and barriers modeled across many \acp{btd}. For our example, we have restricted our analysis to these few events and barriers contained in a single \ac{btd}.

\subsubsection{Conditional Relationships}
\label{sec:evaluation/conditional_relationships}
Here we consider the barrier node B2 in \cref{fig:bowtie-example} as an example for the probability calculation technique described in \cref{subsec:conditional_estimation}. Barrier B2 is part of the event chain $T2 \rightarrow B2 \rightarrow TOP$ and describes the primary control system's ability to recognize when it is approaching a slow-moving or stopped vehicle in our lane of travel (T2) too quickly and slow down before control of the roadway obstruction hazard (H1) is lost. To isolate barrier B2, simulation scenes were generated using our \ac{sdl} where T2 was the only threat condition present but all other system and environmental parameters were free to vary. It is important for these scenes to cover the range of expected system states and environmental conditions since the resulting dataset is used to estimate the conditional probability of success for barrier B2. A total of 300 simulation scenarios were used for estimating the probability of barrier B2. As a convenience for our example, the threat (T2) was set to occur once during each scene, regardless of other state parameters, which allows the denominator of this ratio to be set as $R(T2 | \textbf{s}) = 1$ occurrence per scene. 

As perception \ac{lec} is the primary controller, the success rate of barriers B1 and B2 will be dependent on the image quality. The image quality will in turn be dependent on image blurriness, occlusion, and environmental conditions. While the level of blur and occlusion for each camera is a configurable simulation parameter, our example system is not provided this information and instead relies on blur and occlusion detectors for each of the three cameras as discussed in \cref{subsec:system_setup}. Each detector provides a boolean output indicating if the level of blur or occlusion exceeds a fixed threshold. The primary \ac{lec} is also susceptible to \ac{ood} data, and an assurance monitor was trained for this \ac{lec} to detect such conditions. 

This results in barrier $B2$ being dependent on a total of 7 state-variables described as the set $S^{B2}$. For the 6 boolean variables, Laplace's rule of succession shown in \cref{eqn:rule_of_succession} was applied to the simulation dataset resulting in probability table in the lower-right section of \cref{fig:bowtie-example}. A sigmoid function was chosen to model the conditional relationship with the continuous assurance monitor output. Maximum likelihood estimation was used to produce the function $P(x | \textbf{s}.m.LEC)$ shown in \cref{fig:bowtie-example} where $\textbf{s}.m.LEC$ represents the output of the \ac{lec} assurance monitor. Each of these single variable functions was combined into the multivariate conditional probability distribution $f_b(B2, \textbf{s})$ using \cref{eqn:multivariate_dist}. Note that the \ac{lec} was observed to be similarly effective in identifying pedestrians and vehicles, and a simplifying assumption was made to use the same conditional probability function for both barriers B1 and B2 given by function $f_b$(x = (B1,B2), \textbf{s}) in \cref{fig:bowtie-example}.

The \ac{aebs} described by barrier B3 is independent of the camera images and relies on the forward looking radar. The level of noise in our simulated radar sensor increased with increasing precipitation levels, indicating that the effectiveness of barrier B3 would likely decrease as precipitation increased. Also, failure of the radar sensor may occur on a random basis which reduces the effectiveness of the \ac{aebs} to zero. A similar conditional estimation process as used for B2 was applied here to calculate the function $f_b(B3, \textbf{s})$ shown in \cref{fig:bowtie-example}.

\subsubsection{Results}
\label{sec:results}
To validate the \ac{resonate} framework, the \ac{av} was tasked to navigate 46 different validation scenes that were generated using our \ac{sdl}. In these scenes, the weather\_description parameters of cloud (c), precipitation (p), and precipitation deposits (d) were varied in the range [0,100]. Other adversities were synthetically introduced using OpenCV including increased image brightness, camera occlusion (15\%-30\% black pixels), and camera blur (using 10x10 Gaussian filters). During each simulation, the ReSonAte's risk calculations continuously estimate the hazard (or collision) rate $h(t)$ based on changing environmental conditions, presence of faults, and outputs from the runtime monitors. The frequency of the risk estimation can be selected either based on the vehicle's speed, environmental changes, or available compute resources. In our experiments, we estimate the risk every inference cycle.

\begin{figure}[t]
 \centering
 \includegraphics[width=\columnwidth]{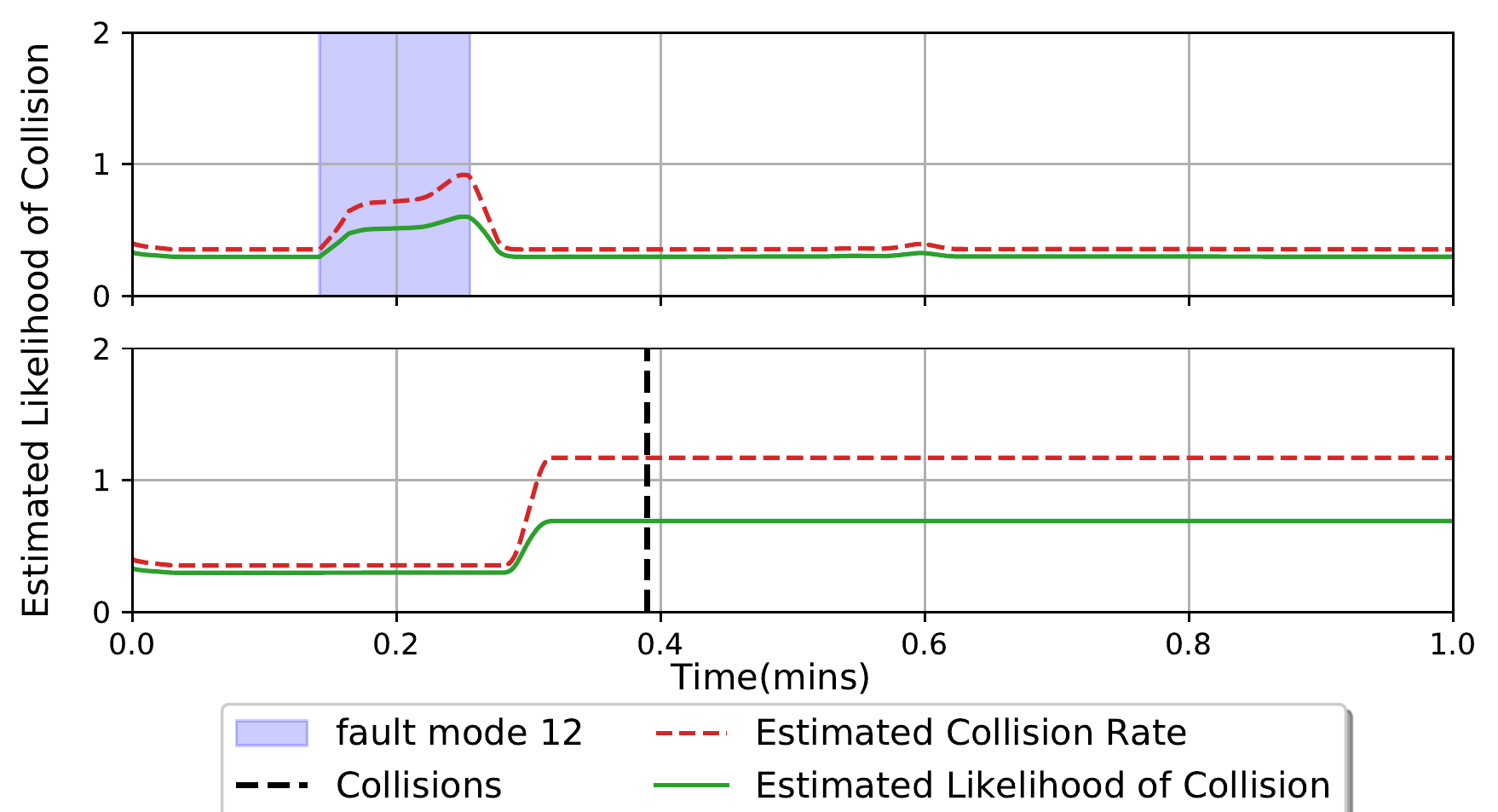}
 \caption{ReSonAte estimated collision rate and likelihood of collision for 2 validation scenes. (Top) Scene1 - nominal scene with good weather and an intermittent occlusion fault for left and center cameras. (Bottom) Scene2 - initially nominal scene until 27 seconds when image brightness is increased. A collision occurs at 38 seconds denoted by the vertical dotted line.}
 \label{fig:scene-risk}
 \vspace{-0.15in}
\end{figure}

\cref{fig:scene-risk} shows the estimated collision rate and the likelihood of collision as the \ac{av} navigates 2 validation scenes. The occurrence of a collision can be described as a random variable following a Poisson distribution where the estimated collision rate is the expected value $\lambda$. The likelihood of collision can then be computed as ($1-e^{-\lambda \cdot t}$), where $\lambda$ is the estimated collision rate and $t$ is the operation time which is fixed to 1 minute in our experiments. 



\begin{figure}[t]
 \centering
 \includegraphics[width=\columnwidth]{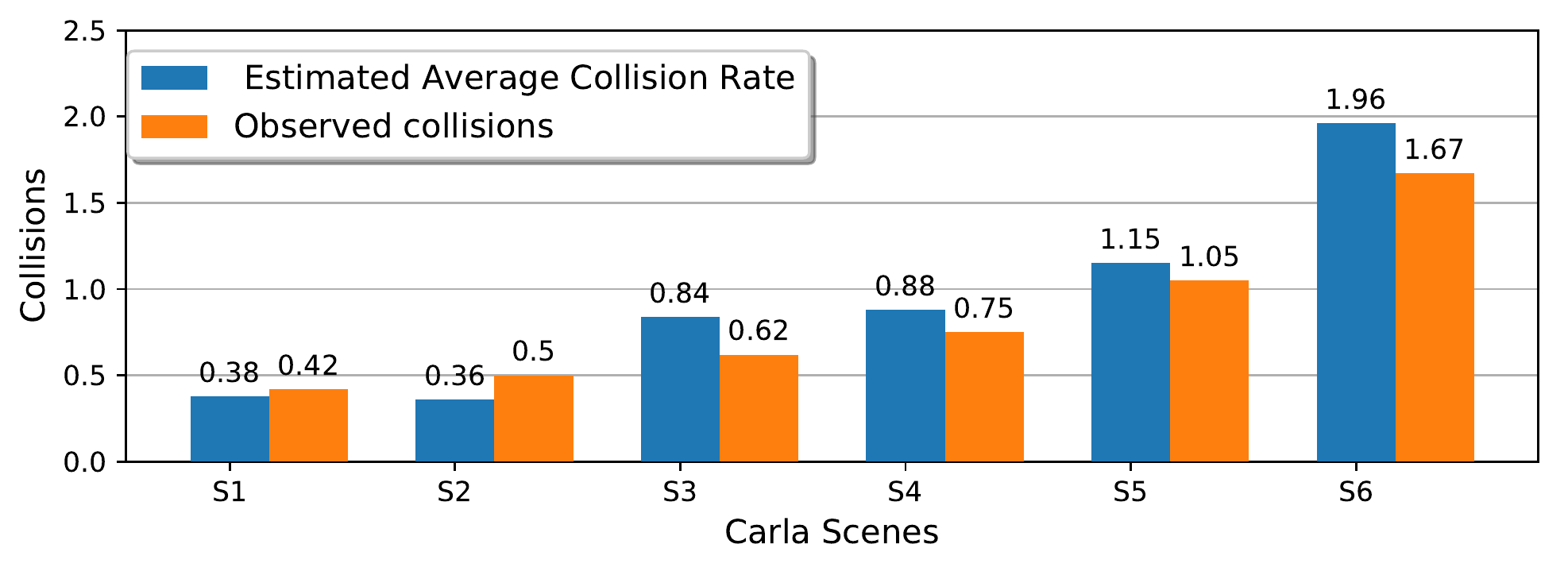}
 \caption{ReSonAte estimated average collision rate vs. observed collisions compared across 6 validation scenes. The results are averaged across 20 simulation runs for each scene. Each subsequent scene represents increasingly adverse weather and component failure conditions.}
 \label{fig:risk-across-scenes}
\vspace{-0.15in}
\end{figure}


\cref{fig:risk-across-scenes} shows the estimated average collision rate plotted against the observed collisions for 6 validation scenes described in the figure caption. The estimated average collision rate is calculated as $\frac{\int_{T_1}^{T_2} h(t) \,dt}{T_2-T_1}$, where, h(t) is the estimated collision rate, $T_1$ = 0 and $T_2$ = 1 minute for our simulations. A moving average is used to smooth the estimated collision rate and a window size of 20 was selected to balance the desired smoothing against the delay incurred by the moving average. An overall trend in the plot shows a strong correlation between the actual and the estimated collisions. Also, a visible trend is that the collision rate changes with the weather patterns, increasing for adverse weather conditions (S5-S6). However, the estimated risk tends to slightly overestimate when the collision rate is greater than 1.0.



\begin{figure}[t]
    \begin{subfigure}{.49\columnwidth}
      \centering
      \includegraphics[width=\linewidth]{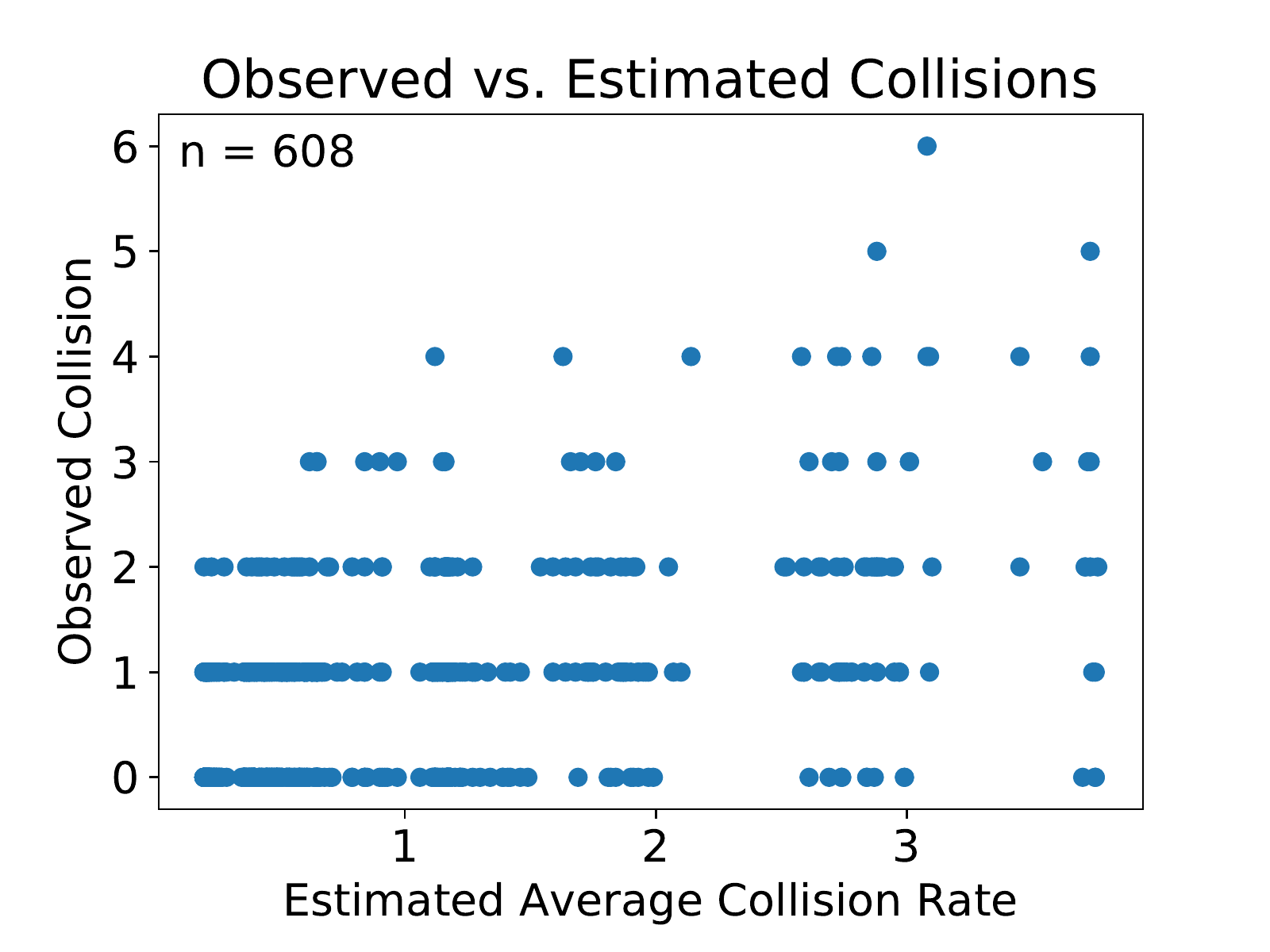}
      \caption{Validation data scatter plot.}
      \label{fig:results_scatterplot}
    \end{subfigure}
    \begin{subfigure}{.49\columnwidth}
      \centering
      \includegraphics[width=\linewidth]{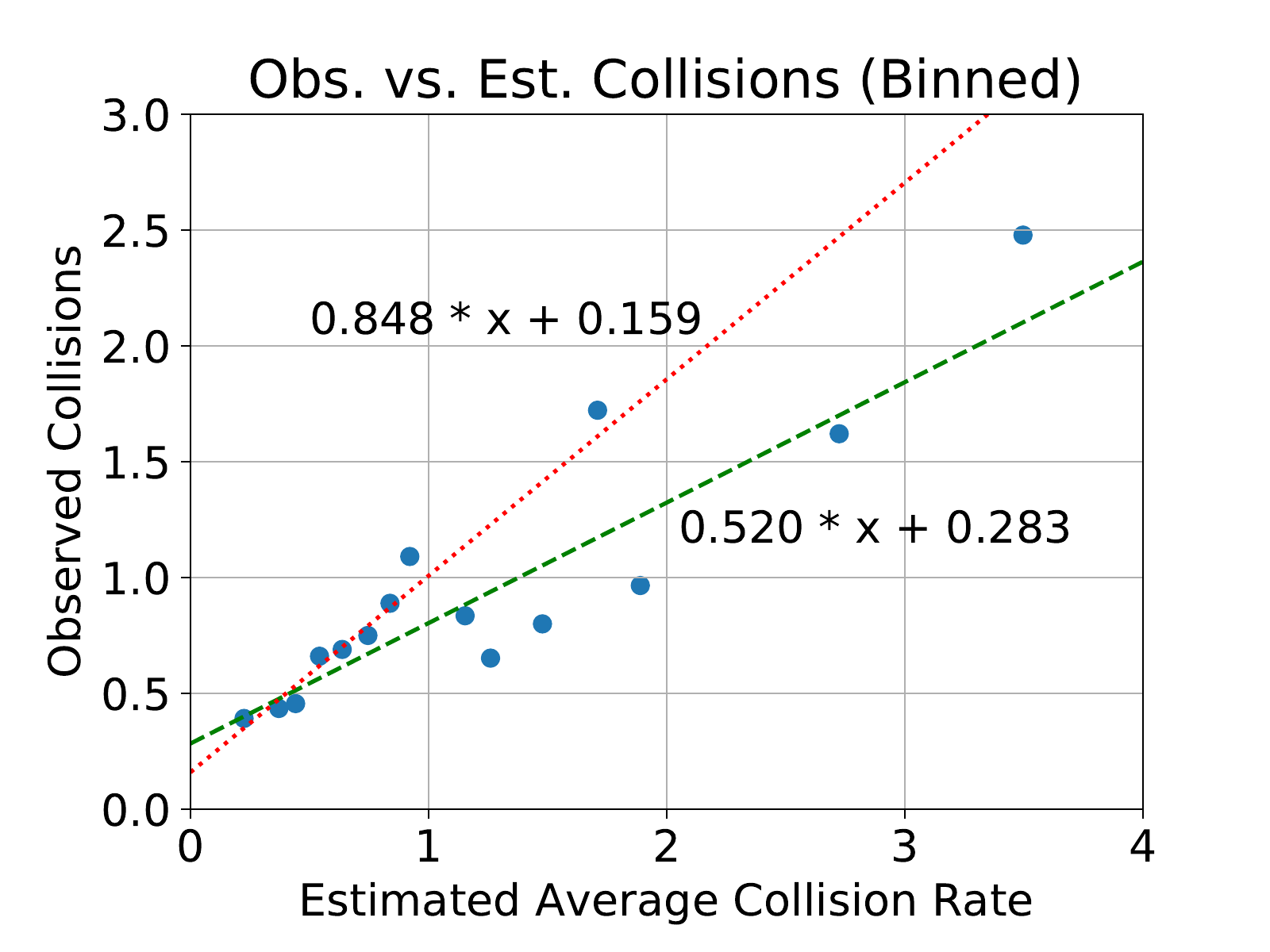}
      \caption{Binned validation data.}
      \label{fig:results_binned}
    \end{subfigure}
 \caption{Results from validation scenes for the \ac{resonate} framework. Each data point shown in \cref{fig:results_scatterplot} represents the outcome of one simulated scene with the actual number of collisions observed plotted against the estimated average collision rate. These same data points have been divided into bins and averaged in \cref{fig:results_binned}, along with two least-squares fit trend lines.}
 \label{fig:validation_results}
\vspace{-0.1in}
\end{figure}

Further, each of the 608 data points shown in \cref{fig:results_scatterplot} represents the outcome of one simulated scene with the actual number of collisions plotted against the average collision rate estimated by \ac{resonate}. Since the occurrence of a collision is a probabilistic event, there is significant variation in the actual number of collisions observed in each scene. Using our dynamic rate calculation approach, the rate parameter $\lambda$ of the Poisson distribution changes for each scene as shown on the x-axis of \cref{fig:results_scatterplot}. Maximum likelihood analysis was used to compare our dynamic approach against a static, design-time collision rate estimate where $\lambda$ is fixed for all scenes. The observed average collision rate across all scenes was found to be 0.829 collisions per minute. Using this static $\lambda$ value gave a log likelihood of -740.7 while the dynamically updated $\lambda$ resulted in a log likelihood of -709.8 for a likelihood ratio of 30.9 in favor of our dynamic approach. Note that the static collision rate used here is a posterior estimate calculated from the true number of collisions observed. Any static risk estimate made without this post facto knowledge would result in a lower likelihood value and further increase the gap between the dynamic and static approaches.

To better show the correlation between estimated and actual collisions, the same data points have been divided into bins and averaged in \cref{fig:results_binned} along with two least-squares fit trend lines.  The dashed green trend line shows a linear fit to the complete data set while the dotted red line shows a linear fit to only those data points where the average estimated collision rate was less than or equal to 1.0. Both trend lines show a strong positive correlation between the estimated collision rate and the number of observed collisions, but the dotted red line more closely resembles the desired 1-to-1 correspondence between estimated and actual collisions (i.e. slope of 1). These results indicate that our dynamic risk calculation tends to over-estimate when the estimated collision rate is greater than 1.0. 

\input{resource-table}

\cref{Table:resource} shows the resource requirements and execution times for system with different configurations. As seen from the shaded columns, the additional sensors, and system monitors, particularly the resource intensive $\beta$-VAE monitor, increases the GPU memory used by $\sim$ 40\% and execution time by $\sim$ 0.2 seconds. However, the ReSonAte risk calculations require minimal computational resources taking only 0.3 milliseconds.

\subsection{Unmanned Underwater Vehicle}
In this section we discuss the primary results of applying ReSonAte to a \ac{uuv} testbed based on the BlueROV2 \cite{robotics2016bluerov2} vehicle. The testbed is built on the ROS middleware \cite{quigley2009ros} and simulated using the Gazebo simulator environment \cite{koenig2004design} with the \ac{uuv} Simulator \cite{manhaes2016uuvsimulator} extensions.

\subsubsection{System Overview and \ac{btd} Modeling}
In this example, the \ac{uuv} was tasked to track a pipeline while avoiding static obstacles (e.g., plants, rocks, etc.) as shown in \cref{fig:bluerov_screenshot}. The \ac{uuv} is equipped with 6 thrusters, a forward looking sonar (FLS), 2 side looking sonars (SLS), an IMU, a GPS, an altimeter, an odometer, and a pressure sensor. The vehicle has several ROS nodes for performing pipe tracking, obstacle avoidance, degradation detection, contingency planning, and control. Additional contingency management features such as thruster reallocation, return-to-home, and resurfacing are available. The \ac{uuv} uses the SLS along with the FLS, odometry, and altimeter to generate the HSD commands for pipe tracking and obstacle avoidance. The degradation detector is an \ac{lec} which employs a feed-forward neural network to detect possible thruster degradation based on thruster efficiency information. This \ac{lec} sends information to the contingency manager including the identifier of the degraded thruster, level of degradation, and a value measuring confidence in the predictions. The contingency manager may then perform a thruster reallocation (i.e. adjustment of the vehicle control law) if necessary to adapt to any degradation. A ROS node implementation of \ac{resonate} was used to dynamically estimate and publish the likelihood of a collision.


\subsubsection{Hazard Analysis and \ac{btd} Modeling}
\label{sec:hazard_uuv}
Using the system requirements and its operating conditions we identified \ac{uuv} operating in presence of static obstacles as the hazard condition which could result in the \ac{uuv}'s collision. The static obstacle which appears at a distance less than the desired separation distance of 30 meters is considered to be a threat in this example. Further, the static obstacle appearing at a distance less than a minimum separation distance of 5 meters is considered to be a TOP event for the \ac{btd}. This information was used to outline a \ac{btd} for the \ac{uuv} example. 

\subsubsection{Conditional Relationships}
The probability estimation method described in \cref{subsec:conditional_estimation} was used to compute the conditional probabilities for the \ac{btd}. We used data from 350 simulation scenarios for the conditional probability calculations. These simulations were generated using several scenes generated by varying parameters such as the obstacle sizes (cubes with lengths (0.5,1,2,5,10) meters), the obstacle spawn distance (value in range [5-30] meters), and random thruster failures that were synthetically introduced by varying the thruster1 efficiency in the range [0-60].

\begin{figure}[t]
 \centering
 \includegraphics[width=\columnwidth]{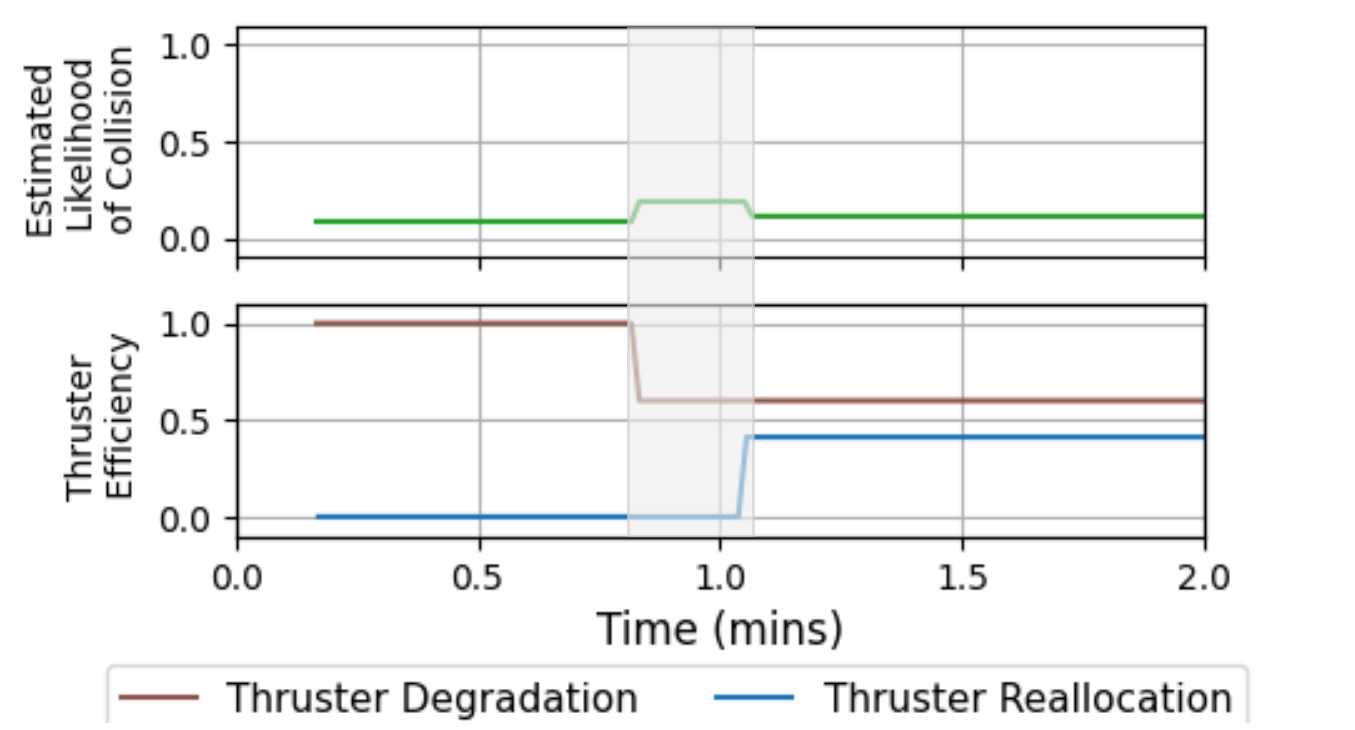}
 \caption{ReSonAte estimated likelihood of collision for the BlueROV2 example. As seen in gray shaded region, the likelihood of collision increases when thruster degradation occurs, then reduces after thruster control reallocation.}
 \label{fig:bluerov-risk}
 \vspace{-0.2in}
\end{figure}

\subsubsection{Results}
\cref{fig:bluerov-risk} shows the ReSonAte estimated likelihood of collision. The efficiency of thruster 1 degrades to 60\% at 45 seconds, and the estimated likelihood of collision increases to 0.25. Soon after, the contingency manager performs a thruster reallocation to improve the \ac{uuv}'s stability, the likelihood of collision decreases to 0.15. We are currently validating the estimated likelihood across large simulation runs.

\vspace{-0.05in}

%% file: resource-table.tex
\begin{table}[t!]
\centering
\renewcommand{\arraystretch}{1.1}
\footnotesize
\begin{tabular}{|c|c|c|c|c|c|}
\hline
                                                                                           & \multicolumn{2}{c|}{\textbf{GPU}}                                                                                          & \multicolumn{2}{c|}{\textbf{CPU}}                                                                                          &                                                                                                   \\ \cline{2-5}
\multirow{-2}{*}{\textbf{\begin{tabular}[c]{@{}c@{}}System\\ Configuration\end{tabular}}} & \textbf{\begin{tabular}[c]{@{}c@{}}Util\\ (\%)\end{tabular}} & \textbf{\begin{tabular}[c]{@{}c@{}}Mem\\ (\%)\end{tabular}} & \textbf{\begin{tabular}[c]{@{}c@{}}Util\\ (\%)\end{tabular}} & \textbf{\begin{tabular}[c]{@{}c@{}}Mem\\ (\%)\end{tabular}} & \multirow{-2}{*}{\textbf{\begin{tabular}[c]{@{}c@{}}Execution\\ Time (s)\end{tabular}}} \\ \hline
\textbf{LEC only}                                                                   & 14.3                                                         & \cellcolor[HTML]{C0C0C0}47.2                                & 17.6                                                         & 10.4                                                        & \cellcolor[HTML]{C0C0C0}0.024                                                                     \\ \hline
\textbf{\begin{tabular}[c]{@{}c@{}}LEC +\\ Runtime Monitors\end{tabular}}      & 14.7                                                         & \cellcolor[HTML]{C0C0C0}86.5                                & 18.5                                                         & 10.5                                                        & \cellcolor[HTML]{C0C0C0}0.217                                                                     \\ \hline
\textbf{\begin{tabular}[c]{@{}c@{}} LEC +\ Monitors + \\ ReSonAte\end{tabular}}             & 16.1                                                         & \cellcolor[HTML]{C0C0C0}87.0                                & 18.7                                                         & 10.4                                                        & \cellcolor[HTML]{C0C0C0}0.218                                                                     \\ \hline
\end{tabular}
\caption{Resource requirements and execution times for different configurations of the system. Average values computed across 20 simulation runs.}
\label{Table:resource}
\vspace{-0.15in}
\end{table}